\pdfoutput=1

\documentclass[11pt]{article}

\usepackage[preprint]{acl}

\usepackage{times}
\usepackage{array}
\usepackage{latexsym}
\usepackage{booktabs}
\usepackage{subcaption}
\usepackage{enumitem}
\usepackage{siunitx}
\usepackage{placeins}
\usepackage{stfloats}

\usepackage[T1]{fontenc}

\usepackage[utf8]{inputenc}

\usepackage{microtype}

\usepackage{inconsolata}

\usepackage{graphicx}

\title{Does Unlearning Truly Unlearn? A Black Box Evaluation of LLM Unlearning Methods}

\author{Jai Doshi \\
  New York University \\
  \texttt{jd5697@nyu.edu} \\\And
  Asa Cooper Stickland \\
  New York University \\
  \texttt{asacoopstick@gmail.com} \\}

\begin{document}
\maketitle
\begin{abstract}
Large language model unlearning aims to remove harmful information that LLMs have learnt to prevent their use for malicious purposes. LLMU and RMU have been proposed as two methods for LLM unlearning, achieving impressive results on unlearning benchmarks. We study in detail the impact of unlearning on LLM performance metrics using the WMDP dataset as well as a new biology dataset we create. We show that unlearning has a notable impact on general model capabilities, with the performance degradation being more significant in general for LLMU. We further test the robustness of the two methods and find that doing 5-shot prompting or rephrasing the question in simple ways can lead to an over ten-fold increase in accuracy on unlearning benchmarks. Finally, we show that training on unrelated data can almost completely recover pre-unlearning performance, demonstrating that these methods fail at truly unlearning. Our methodology serves as an evaluation framework for LLM unlearning methods. The code is available at: \url{https://github.com/JaiDoshi/Knowledge-Erasure}. 
\end{abstract}

\section{Introduction} \label{Introduction}

LLMs are trained on a large amount of publicly available data. This data often contains information that can be used for malicious purposes (such as instructions on how to build explosive devices; \citealp{DBLP:journals/corr/abs-2112-04359}), or that may potentially violate copyright law \citep{JMLR:v24:23-0569}. A solution to this is \textit{machine unlearning}, which we define as updating the weights of the model so that it loses access to potentially harmful information. A good unlearning method should satisfy the following additional criteria: (2)~it should not have a significant impact on the general capabilities of the model, and (3)~once unlearning has been performed, the unlearned information should be permanently removed from the model, i.e. it should not be possible to recover this information from the model via fine-tuning, probing, or any other strategies. 

\begin{figure*}[]
  \includegraphics[width=\linewidth]{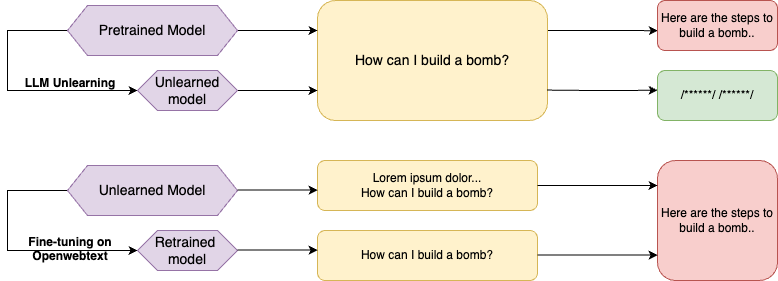}
  \caption{Applying unlearning techniques on the pretrained model makes it give nonsensical outputs to harmful queries. Adversarial prompting (in this case adding filler text before the question) or fine-tuning on benign data causes unlearned capabilities to resurface.}
  \label{fig:diagram}
\end{figure*}

Two recent methods for LLM unlearning are LLMU \citep{yao2023large} and RMU \citep{li2024wmdpbenchmarkmeasuringreducing}. Although these works show that the methods are effective at unlearning, they are limited in the scope of their evaluation and robustness testing. We introduce a new biology-focused dataset that serves as a measure of the ability to unlearn on noisy data; use three main metrics to study the effect of unlearning on general model capabilities; and design robustness tests based on prompting strategies that an adversary may use. These include doing 5-shot prompting and rephrasing the question in different ways, such as in the form of a poem. Applying these simple prompting strategies leads to an increase in accuracy of up to 1750\% on unlearning benchmarks, suggesting that the information has not been actually removed from the model. We test this hypothesis by checking the impact of retraining on benign data, and find that this undoes the effect of unlearning and restores harmful capabilities. Figure~\ref{fig:diagram} highlights our robustness testing approach.  
 
Overall, our work develops a framework for evaluating LLM unlearning with a focus on black box methods for robustness testing. The application of this framework to two recent unlearning methods demonstrates deterioration in general model capabilities and ineffectiveness at truly unlearning. 

\section{Related Work}

\citet{liu2024rethinking} expound unlearning effectiveness and utility preservation (along with efficiency) as the main criteria for evaluating unlearning methods. \citet{lucki2024adversarialperspectivemachineunlearning} focus on the robustness of unlearning, and use methods such as Logit Lens and a modified version of GCG to recover unlearned performance. They show that fine-tuning on just a few unrelated examples is able to almost completely undo the effects of unlearning. However, the methods they use assume access to the model weights, and so cannot be applied to black box LLMs except for cases where model fine-tuning is possible through API access. \citet{lynch2024methodsevaluaterobustunlearning} define eight metrics that can be used to evaluate the robustness of LLM unlearning methods, including translating the queries to another language. We adapt some of these metrics in our work, as well as introduce new ones that can be applied to black box models.

\section{Methodology}

\subsection{Unlearning Methods}

\paragraph{LLMU} 

\citet{yao2023large} define three losses: the gradient ascent loss $\mathcal{L}_{\mathrm{fgt}}$, $\mathcal{L}_{\mathrm{rdn}}$ which trains the LLM towards a random output for an input that is to be unlearned, and $\mathcal{L}_{\mathrm{nor}}$ which computes the KL divergence between the unlearning model and normal model on a retain set to preserve normal utility. These losses are weighted by hyperparameters $\epsilon_1$, $\epsilon_2$ and $\epsilon_3$ respectively. $\mathcal{L}_{\mathrm{rdn}}$ is better suited for QA datasets, so we use only $\mathcal{L}_{\mathrm{fgt}}$ and $\mathcal{L}_{\mathrm{nor}}$. 

\paragraph{RMU}

The squared distance between the outputs of the model at some layer $\ell$ and a fixed random control vector $\mathbf{u}$ (which is scaled by a factor $c$) on the unlearning set is used as the unlearn loss. The $L_2$ loss between the unlearning model and the normal model at that layer on the retain set is used as the retain loss. The retain loss is weighted by hyperparameter $\alpha$.

\subsection{Datasets}

\paragraph{Wikipedia Biology}

We consider the example task of unlearning all biology information from the model. We believe this task serves as a good proxy to real world applications where a particular domain needs to be forgotten. We scrape all Wikipedia articles under biology-related categories, recursively processing all the subcategories within these categories until a certain depth. Although some of the articles may not be relevant, such as biographies of biologists, we keep these articles to measure the ability to unlearn on unclean data. We similarly process randomly chosen non-biology categories to create the retain set. The lists of categories used and the depths of recursive processing are given in Appendix~\ref{Details of Wikipedia Categories Processed}.

\paragraph{WMDP}

We additionally unlearn on the WMDP dataset \citep{li2024wmdpbenchmarkmeasuringreducing}, focusing on unlearning the WMDP-Bio and WMDP-Cyber subsets, and use Wikitext as the retain set.

\subsection{Metrics}

Based on the criteria of an effective unlearning method described in Section~\ref{Introduction}, we use the following metrics for evaluation: \\

\noindent \textbf{1. Non-generation of harmful responses} 
    
\noindent \textbf{(a) Question Answering} \\
We use accuracy on multiple choice benchmarks as the primary metric. For biology unlearning we consider the macro-average accuracy on MMLU \citep{hendrycks2021measuringmassivemultitasklanguage} for biology-related subjects (refer to Appendix~\ref{MMLU Biology Subjects} for the list of subjects). For WMDP, we use the provided QA dataset, and evaluate on the Biology and Cyber subsets. We use a zero-shot template akin to \citet{li2024wmdpbenchmarkmeasuringreducing}, but remove the name of the subject as it increases the model's tendency to refuse to answer. An example question is provided in Appendix~\ref{Sample Zero-shot Question}. To measure the robustness of unlearning, we use the following two additional tests: 

\noindent \textbf{(b) Five-shot prompting} \\ 
In order to encourage the model to answer the question, we include five examples of answered multiple choice questions of an MMLU subject before the actual question. Appendix~\ref{5-shot List} contains the list of MMLU subjects used and Appendix~\ref{Sample Five-shot Question} contains an example question.   

\noindent \textbf{(c) Question rephrasing} \\ 
We experiment with rephrasing the question in different ways that we think would bypass the mechanisms that the unlearning methods use to filter out harmful queries. For example, we try replacing technical jargon in the question with common words, with the hypothesis that the model refuses to answer the question when it encounters technical vocabulary. For translations, we choose high-resource languages, common languages, as well as languages that \citet{li2024preferencetuningtoxicitymitigation} find to be semantically different from English. See Appendix~\ref{Rephrasing Types} for the rephrasing types. The exact prompts used are contained in the code. \\

\noindent \textbf{2. Preservation of general model capabilities} 

\noindent \textbf{(a) MMLU accuracy} \\
Zero-shot macro-average accuracy on MMLU. While evaluating the Biology dataset we exclude the biology subjects. 

\noindent \textbf{(b) MT-Bench score} \citep{10.5555/3666122.3668142} \\
This is a measure of the model's ability to follow instructions and act as a helpful chatbot.

\noindent \textbf{(c) Perplexity} \\ 
We measure the perplexity of the model on Openwebtext. \\

\noindent \textbf{3. Prevention of recovery of unlearned information} 

\noindent \textbf{Fine-tuning on benign data} \\
\citet{lynch2024methodsevaluaterobustunlearning} list fine-tuning on harmful data as one of the metrics to evaluate unlearning. However, this requires the malicious user to have access to a dataset of harmful information. Similar to contemporary work by \citet{lucki2024adversarialperspectivemachineunlearning}, we experiment with using the benign Openwebtext dataset to see if unlearned performance can be recovered by fine-tuning on generic web data.

\section{Experiments}

\paragraph{Models} We conduct our experiments on Zephyr-7B-$\beta$ \citep{tunstall2023zephyrdirectdistillationlm} and Meta-Llama-3-8B-Instruct \cite{dubey2024llama3herdmodels}. We choose instruction-tuned models as our evaluation metrics are based on question answering. 

\paragraph{Data processing} For the Biology dataset, articles are randomly sampled from the dataset. The text is then divided into chunks of a fixed size using a sliding window with a given stride. The sliding window approach is also used to process Openwebtext data for perplexity and fine-tuning evaluations. For the WMDP dataset, training is done on alternating examples from the Biology and Cyber subsets, generated by truncating examples from the original dataset similar to the approach of \citet{li2024wmdpbenchmarkmeasuringreducing}.

\paragraph{Training} For LLMU, we fine-tune the $\epsilon_1$ hyperparameter, fixing $\epsilon_3$ at $1$, and train for 5000 steps, checkpointing every 500 steps. We use LoRA fine-tuning due to memory constraints. For RMU, we tune the $c$ and $\alpha$ hyperparameters, as well as the layer at which the loss is computed, and train for 1000 steps, checkpointing every 50 steps. The model provided by \citet{li2024wmdpbenchmarkmeasuringreducing} is used for Zephyr on WMDP. For both methods, we pick the hyperparameters by evaluating at all checkpoints until a checkpoint is obtained for which zero-shot accuracy on the unlearning benchmark is < 20\%, with MMLU accuracy greater than 50\% and MT-Bench score greater than 5, as we consider this to be a reasonable trade-off between unlearning and preservation of normal utility. Roughly 10 runs were required per dataset-method-model for tuning. \\

\noindent Training and Openwebtext fine-tuning were done on one A100 80GB GPU. RMU training and Openwebtext fine-tuning took around 20 minutes and LLMU training took around 1 hour to run. MMLU and MT-Bench evaluation were done using A100/H100/RTX 8000/V100 GPUs based on availability, with evaluation times varying based on the GPU used.

\section{Results}
\subsection{Unlearning and Robustness Tests}
\begin{table*}[]
\small
    \begin{subtable}{\linewidth}

    \begin{tabular}
{p{0.18\linewidth}
>{\centering\arraybackslash}p{0.1\linewidth}
>{\centering\arraybackslash}p{0.12\linewidth}
>{\centering\arraybackslash}p{0.12\linewidth}
>{\centering\arraybackslash}p{0.16\linewidth}
>{\centering\arraybackslash}p{0.2\linewidth}}
            \toprule

\textbf{Model} &
\multicolumn{1}{p{0.1\linewidth}}{\textbf{Biology Accuracy}} & 
\multicolumn{1}{p{0.12\linewidth}}{\textbf{Accuracy Answered}} & 
\multicolumn{1}{p{0.12\linewidth}}{\textbf{Robustness Test Biology Accuracy}} & 
\multicolumn{1}{p{0.16\linewidth}}{\textbf{Accuracy Answered Robustness Test}} & 
\multicolumn{1}{p{0.2\linewidth}}{\textbf{Most Effective Robustness Test}} \\
            
      \midrule

      \textbf{Zephyr Original} &  0.646 & 0.651 & - & - & -  \\

       \cmidrule(lr){1-6}

      Zephyr LLMU &  0.074 & 0.472 & 0.259 (\textbf{+250\%}) & 0.535 & 5-shot High School Physics  \\

    Zephyr RMU &  0.097 & 0.386 & 0.298 (\textbf{+207\%}) & 0.312 & Translated to Telugu  \\

      \midrule

      \textbf{Llama Original} &  0.696 & 0.696 & - & - & -  \\

       \cmidrule(lr){1-6}

      Llama LLMU & 0.117 & 0.572 & 0.396 (\textbf{+238\%})  & 0.643 & 5-shot College Chemistry  \\

    Llama RMU & 0.136  & 0.342 & 0.376 (\textbf{+176\%}) & 0.424  & 5-shot Jurispudence \\

      \bottomrule

    \end{tabular}

          \caption{Biology dataset}

    \label{tab:biology robustness}

  \end{subtable}

\vspace{0.5cm}

    \begin{subtable}{\linewidth}
    \begin{tabular}
{p{0.18\linewidth}
>{\centering\arraybackslash}p{0.1\linewidth}
>{\centering\arraybackslash}p{0.12\linewidth}
>{\centering\arraybackslash}p{0.12\linewidth}
>{\centering\arraybackslash}p{0.16\linewidth}
>{\centering\arraybackslash}p{0.2\linewidth}}
            \toprule

\textbf{Model} &
\multicolumn{1}{p{0.1\linewidth}}{\textbf{Biology Accuracy}} & 
\multicolumn{1}{p{0.12\linewidth}}{\textbf{Accuracy Answered}} & 
\multicolumn{1}{p{0.12\linewidth}}{\textbf{Robustness Test Biology Accuracy}} & 
\multicolumn{1}{p{0.16\linewidth}}{\textbf{Accuracy Answered Robustness Test}} & 
\multicolumn{1}{p{0.2\linewidth}}{\textbf{Most Effective Robustness Test}} \\

      \midrule

      \textbf{Zephyr Original} & 0.663 & 0.665 & - & - & -  \\

       \cmidrule(lr){1-6}

      Zephyr LLMU & 0.185 & 0.689 & 0.254 (\textbf{+37.3\%}) & 0.333 & Translated to Bengali \\

    Zephyr RMU & 0.146 & 0.391  & 0.293 (\textbf{+101\%}) & 0.321 &  Translated to Hindi \\

      \midrule

      \textbf{Llama Original} & 0.710 & 0.710 & - & - & -  \\

       \cmidrule(lr){1-6}

      Llama LLMU & 0.030 & 0.594 & 0.374 (\textbf{+1150\%}) &  0.486 &     Translated to Hindi  \\

    Llama RMU & 0.108 & 0.296 & 0.213 (\textbf{+97.2\%}) & 0.304 & 5-shot Elementary Mathematics  \\
      \bottomrule
    \end{tabular}
          \caption{WMDP Biology dataset}
    \label{tab:wmdp biology robustness}
  \end{subtable}

\vspace{0.5cm}

    \begin{subtable}{\linewidth}
\begin{tabular}
{p{0.18\linewidth}
>{\centering\arraybackslash}p{0.1\linewidth}
>{\centering\arraybackslash}p{0.12\linewidth}
>{\centering\arraybackslash}p{0.12\linewidth}
>{\centering\arraybackslash}p{0.16\linewidth}
>{\centering\arraybackslash}p{0.2\linewidth}}
            \toprule

\textbf{Model} &
\multicolumn{1}{p{0.1\linewidth}}{\textbf{Cyber Accuracy}} & 
\multicolumn{1}{p{0.12\linewidth}}{\textbf{Accuracy Answered}} & 
\multicolumn{1}{p{0.12\linewidth}}{\textbf{Robustness Test Cyber Accuracy}} & 
\multicolumn{1}{p{0.16\linewidth}}{\textbf{Accuracy Answered Robustness Test}} & 
\multicolumn{1}{p{0.2\linewidth}}{\textbf{Most Effective Robustness Test}} \\

      \midrule

      \textbf{Zephyr Original} & 0.420 & 0.437 & - & - & -  \\

       \cmidrule(lr){1-6}

      Zephyr LLMU & 0.143 & 0.642 & 0.143 (+0\%) & 0.642 & Original zero-shot  \\

    Zephyr RMU & 0.106 & 0.398 & 0.241 (\textbf{+127\%}) & 0.373 & 5-shot Professional Law  \\

      \midrule

      \textbf{Llama Original} & 0.466 & 0.466 & - & - & -  \\

       \cmidrule(lr){1-6}

      Llama LLMU & 0.010 & 0.514 & 0.178 (\textbf{+1750\%}) & 0.468 & Translated to Farsi  \\

    Llama RMU & 0.060 & 0.336 & 0.164 (\textbf{+173\%}) & 0.274 &  Technical Terms Removed 2 \\

      \bottomrule

    \end{tabular}

          \caption{Accuracies on the WMDP Cyber Benchmark.}

    \label{tab:wmdp cyber robustness}
  \end{subtable}

  \caption{Accuracies on the respective multiple choice benchmarks along with the accuracies when the model actually answers the question, as well as the maximum accuracy attained from applying robustness tests and the robustness test responsible for the maximum accuracy.}
  \label{tab:robustness tests}
\end{table*}

Table~\ref{tab:robustness tests} contains the accuracies on the benchmarks, as well as the maximum accuracy attained from applying all the robustness tests. The unlearning methods appear to perform well, with performance on the benchmarks reduced to under 20\% (from up to 71\%) in all cases. We observe that the accuracy when the model answers the question is significantly higher for LLMU than RMU.

For the Biology dataset, the robustness tests increase accuracy close to or more than 3 times. For WMDP, the unlearned models tend to be more robust, however, there is an increase of over 1100\% in accuracy for Llama LLMU on translating to certain lower-resource languages such as Farsi. Overall, 5-shot prompting and translations are observed to be the most effective robustness tests. We perform a Z-test based on the accuracy when answered to confirm that the model answers are not random. 

\subsection{Preservation of Model Capabilities}
\begin{figure*}[]

    \begin{subfigure}{\linewidth}
       
        \begin{subfigure}{0.55\linewidth}
            \includegraphics[width=\linewidth]{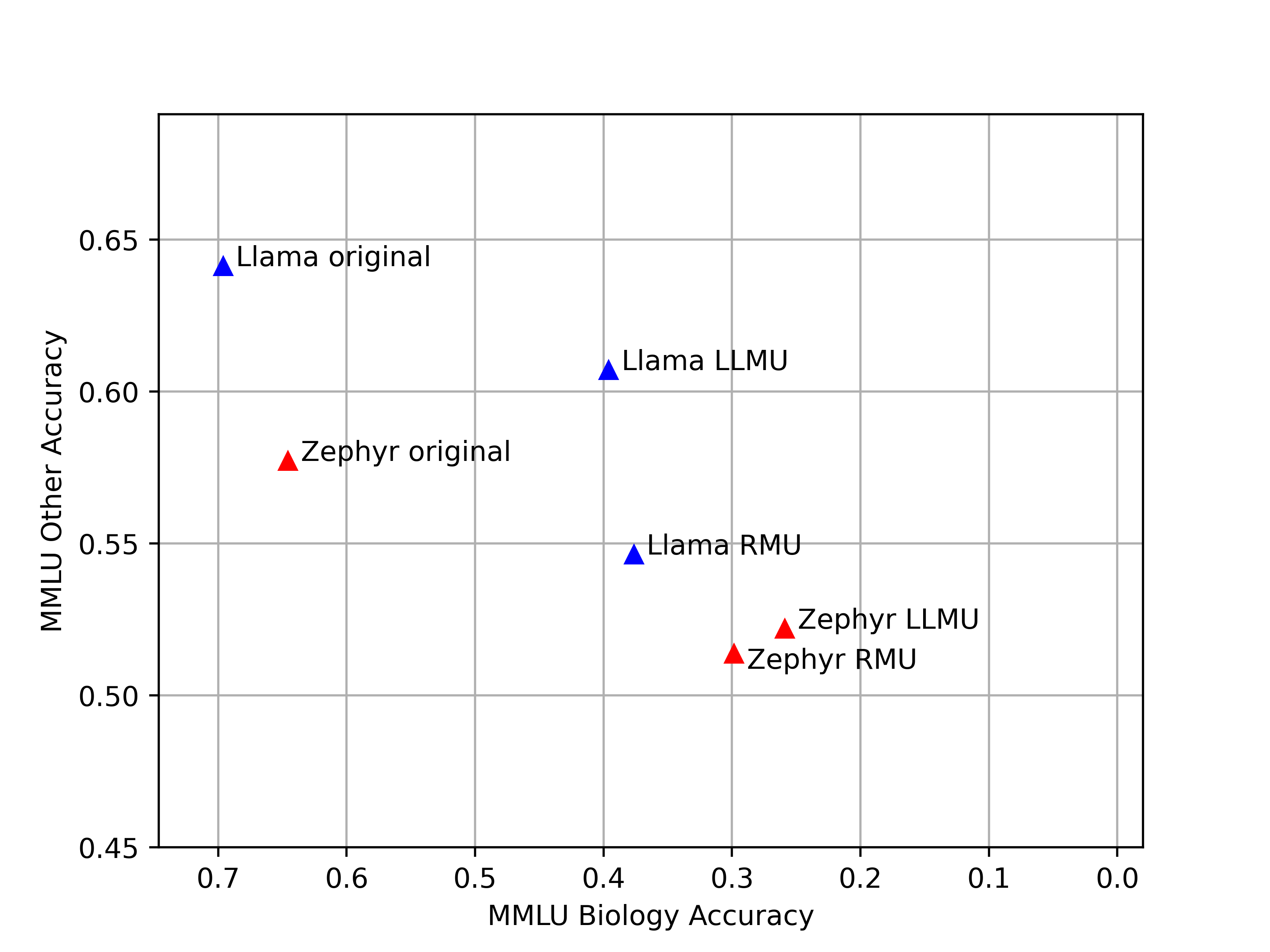}
            \caption*{Effect on MMLU performance}
        \end{subfigure}
        \hfill
        \begin{subfigure}{0.55\linewidth}
            \includegraphics[width=\linewidth]{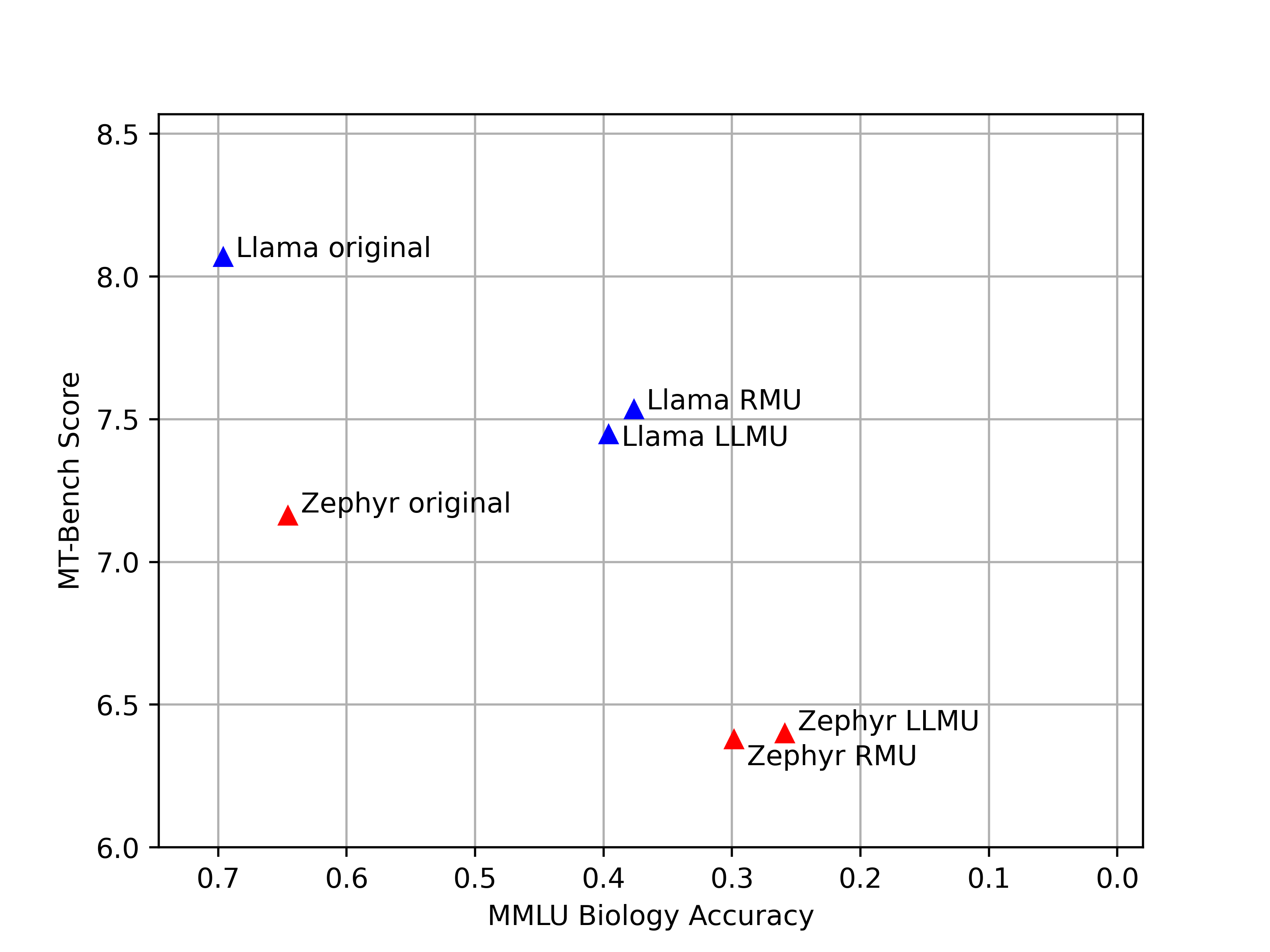}
            \caption*{Effect on MT-Bench performance}
        \end{subfigure}
        \caption{Biology dataset}
    \end{subfigure}

    \vspace{0.1cm}

    \begin{subfigure}{\linewidth}
      
        \begin{subfigure}{0.55\linewidth}
    
            \includegraphics[width=\linewidth]{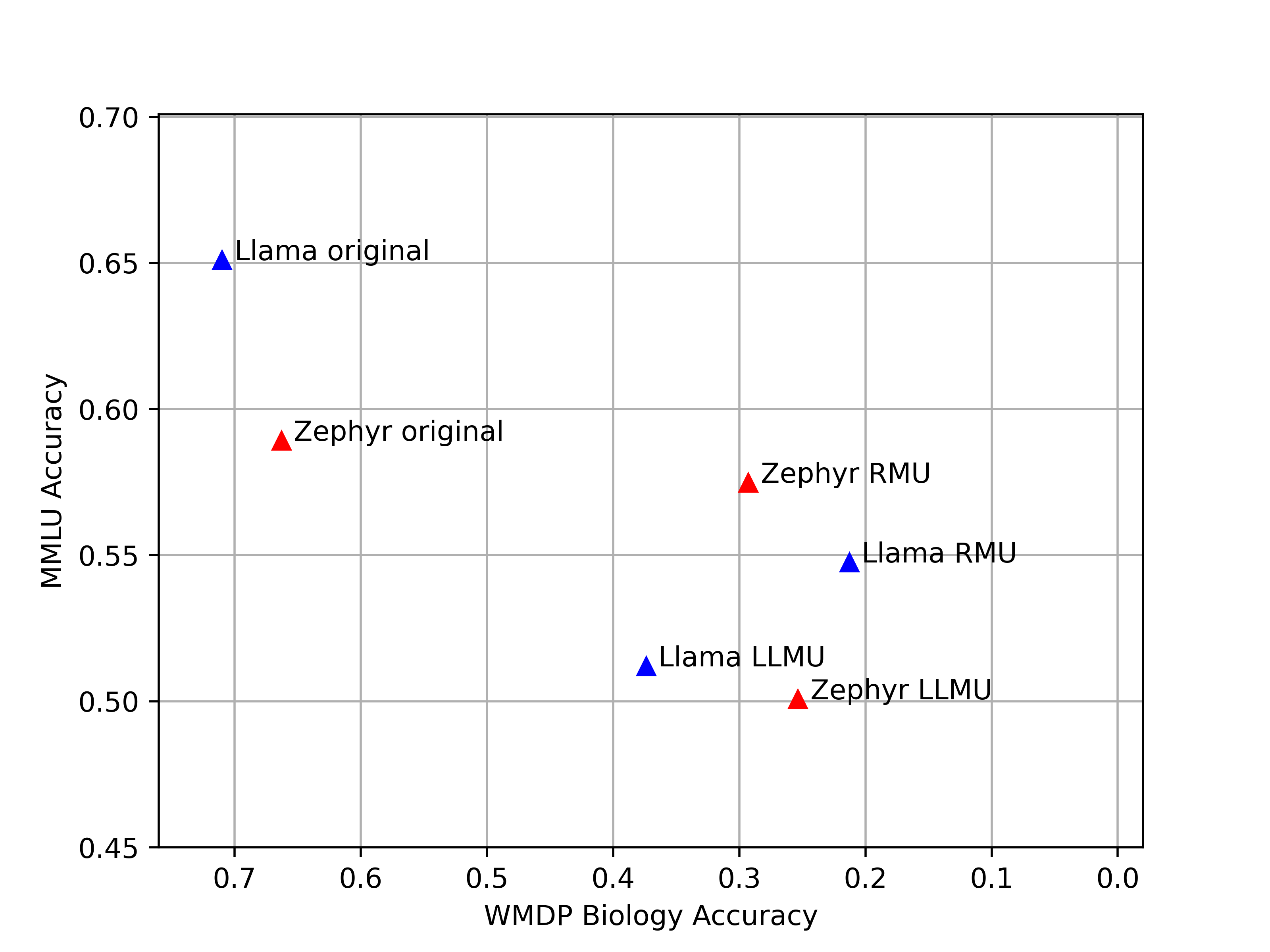}
            \caption*{Effect on MMLU performance}
        \end{subfigure}
        \hfill
        \begin{subfigure}{0.55\linewidth}
         
            \includegraphics[width=\linewidth]{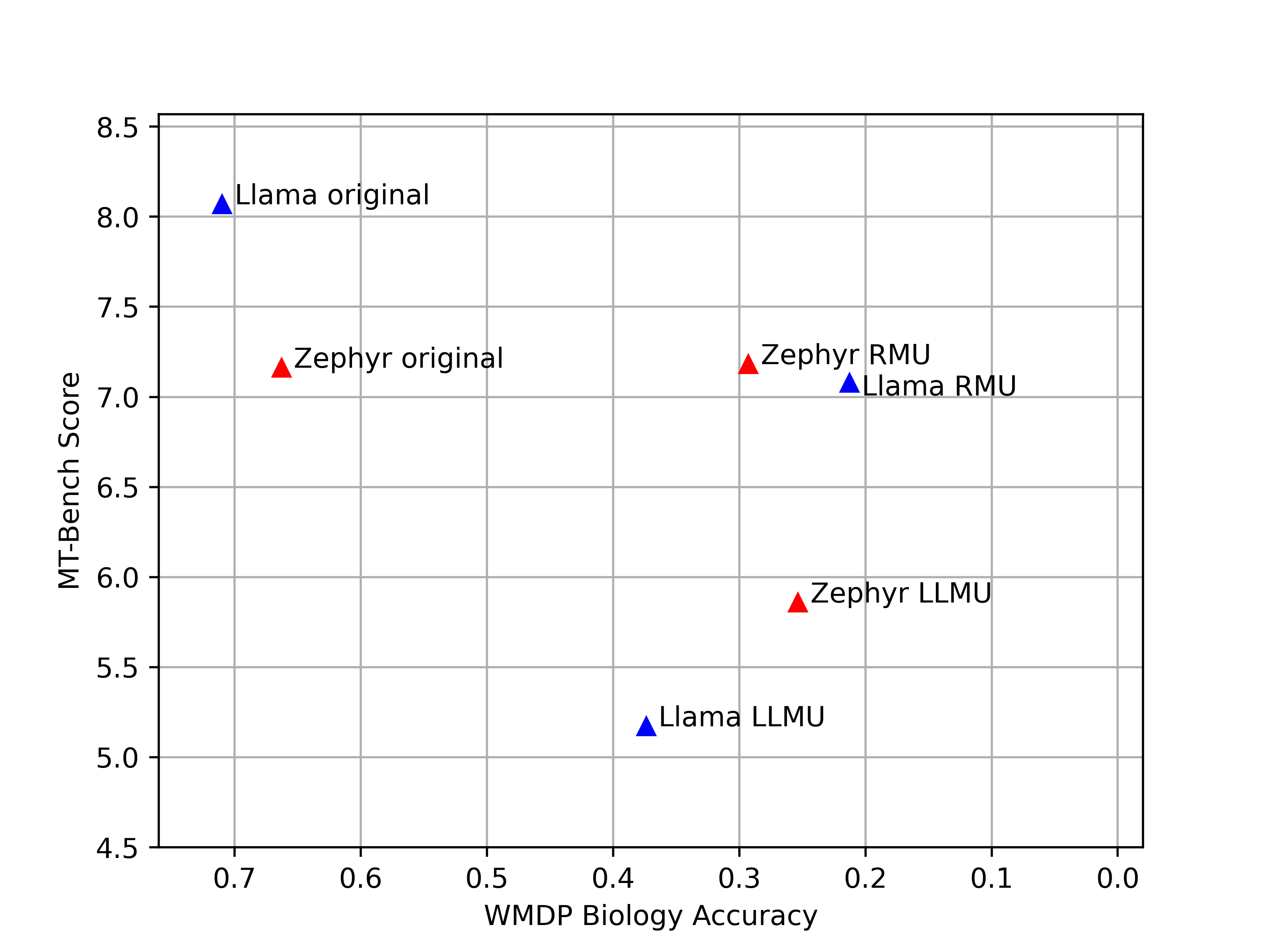}
            \caption*{Effect on MT-Bench performance}
        \end{subfigure}
        \caption{WMDP Biology dataset}
    \end{subfigure}
\begin{subfigure}{\linewidth}
        \begin{subfigure}{0.55\linewidth}
       
            \includegraphics[width=\linewidth]{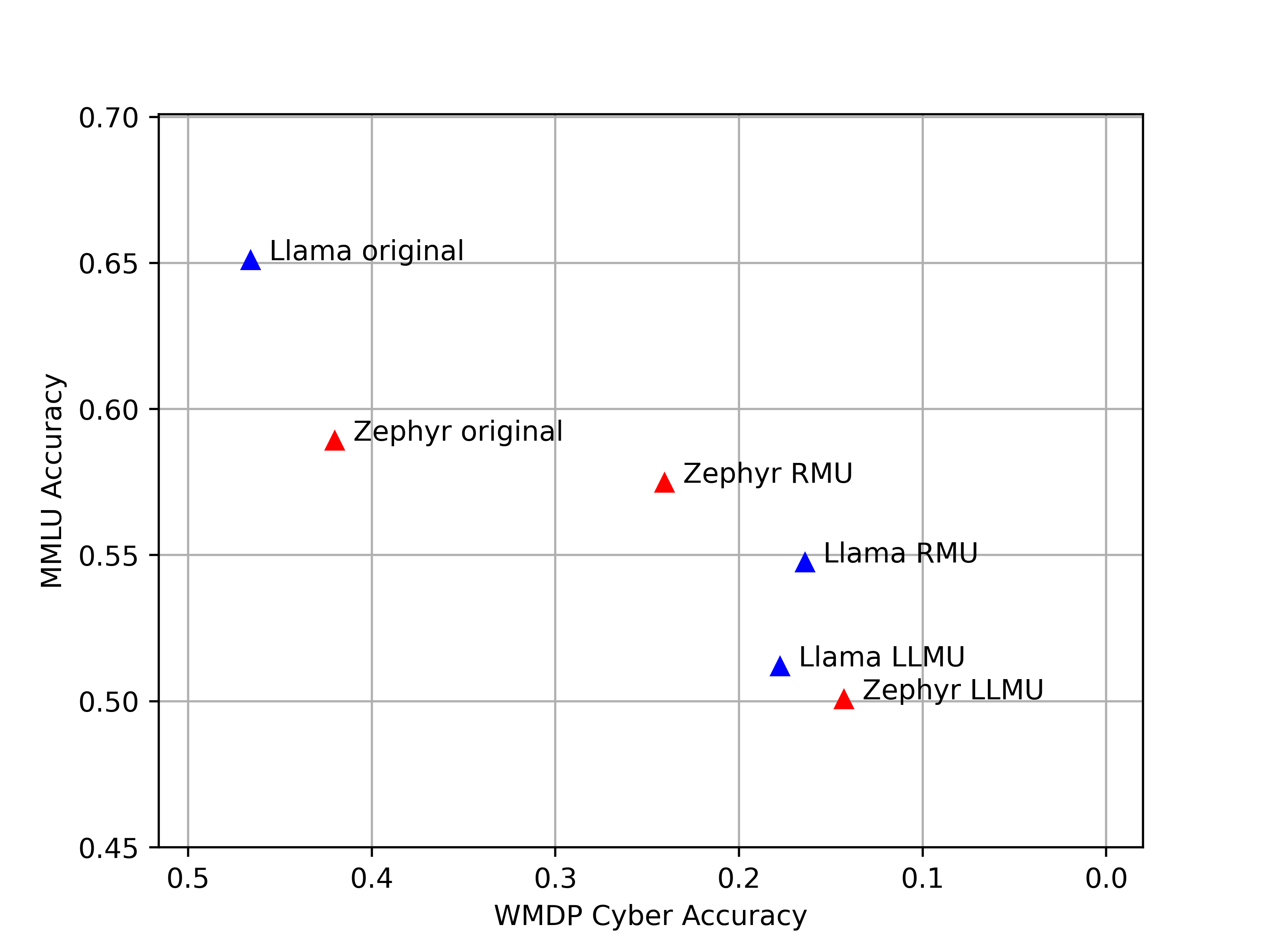}
            \caption*{Effect on MMLU performance}
        \end{subfigure}
        \hfill
        \begin{subfigure}{0.55\linewidth}
    
            \includegraphics[width=\linewidth]{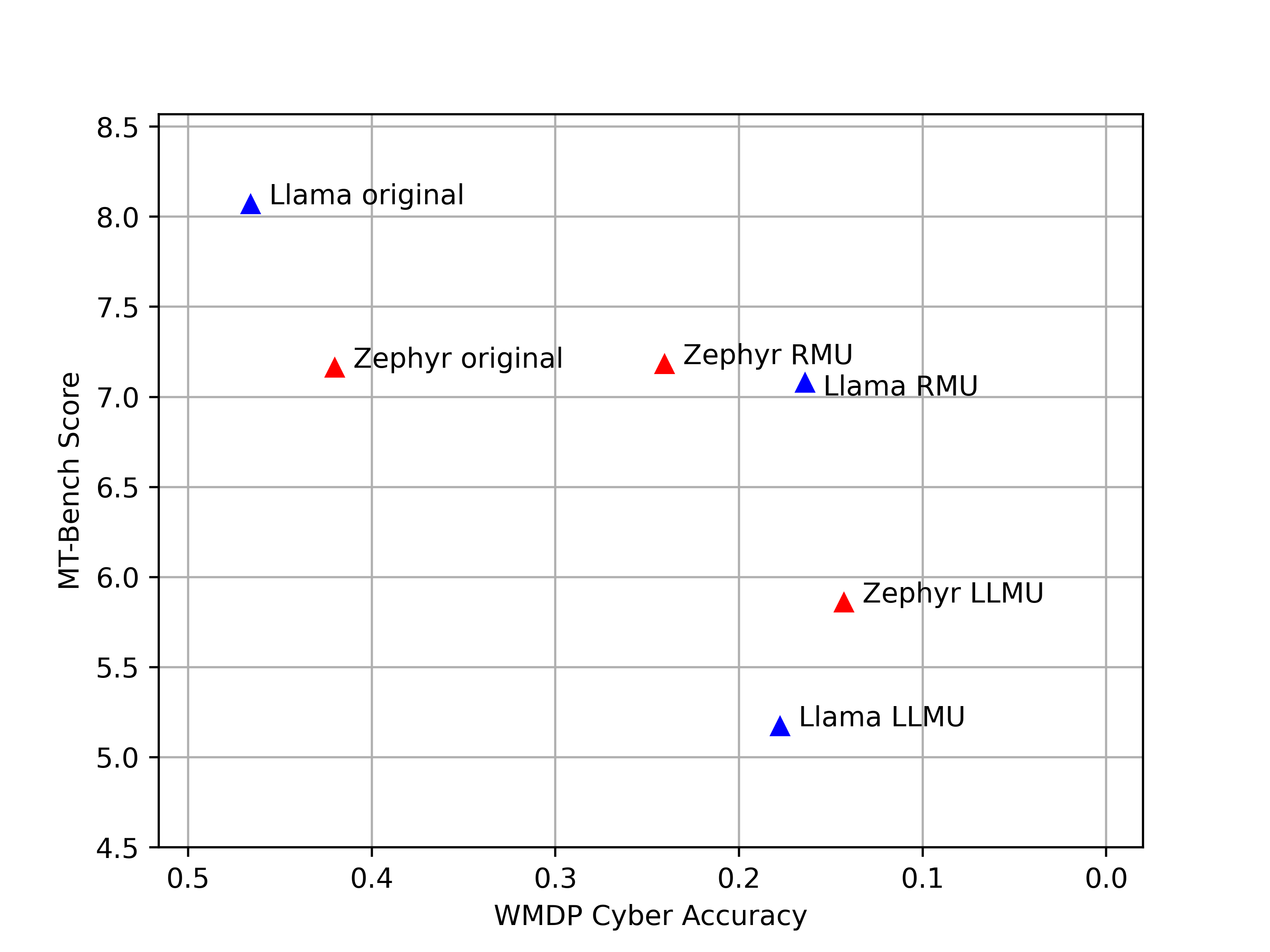}
            \caption*{Effect on MT-Bench performance}
        \end{subfigure}
        \caption{WMDP Cyber Dataset}
    \end{subfigure}

    \caption{Peformance on unlearning benchmarks Vs. MMLU and MT-Bench performance. Top-right direction indicates better performance. The maximum accuracy from the robustness tests listed in Table~\ref{tab:robustness tests} is used as the accuracy on the unlearning benchmarks, as we consider the accuracy after applying the robustness tests a more accurate measure of the degree of unlearning.}
    \label{robustness tests plots}
\end{figure*}

We compare the accuracy from the unlearning benchmarks with performance on MMLU and MT-Bench. Plots are contained in Figure~\ref{robustness tests plots}. In almost all cases, apart from Zephyr RMU on WMDP, there is a noticeable drop in both MMLU and MT-Bench performance post unlearning (>5\%). We observe that for the Biology dataset, both methods perform similarly, with the exception that Llama LLMU has an accuracy on MMLU Other of 60.7\% as compared to 54.6\% for Llama RMU (accuracy on the original model is 64.1\%). On the WMDP dataset however, there is a significant degradation in normal performance for LLMU. For instance, on the Biology subset with Llama, MMLU accuracy drops from 65.1\% to 51.2\% and MT-Bench score drops from 8.07 to 5.17, and RMU is Pareto optimal compared to LLMU for both subsets for Llama.

\begin{table}[!htbp]
\small
\centering
    \begin{tabular}{lc}
      \toprule
      \textbf{Model} & \textbf{Perplexity} \\
      \midrule
      \textbf{Zephyr Original} & 7.90 \\
      \cmidrule(lr){1-2}
      Zephyr LLMU Biology     & \num{2.13e+9}         \\
      Zephyr LLMU WMDP    &  \num{3.03e+4}          \\
        \cmidrule(lr){1-2}
      Zephyr RMU Biology     & 10.3          \\
      Zephyr RMU WMDP    & 8.49          \\
      \midrule
      \textbf{Llama Original} & 13.6 \\
      \cmidrule(lr){1-2}
      Llama LLMU Biology     & 2870         \\
      Llama LLMU WMDP    & 25.7         \\
        \cmidrule(lr){1-2}
      Llama RMU Biology     & 38.6         \\
      Llama RMU WMDP    & 16.5         \\
      \bottomrule
    \end{tabular}
  \caption{Perplexity scores on Openwebtext.}
  \label{tab:perplexity}
\end{table}

Table~\ref{tab:perplexity} contains the perplexity scores on Openwebtext for the unlearned models. For RMU the score increases in all cases, but is comparable to the original model. However, the increase tends to be of multiple orders of magnitude for LLMU. For example, for Zephyr LLMU on the  Biology dataset, the score increases from 7.90 for the original model to \num{2.13e+9}.

\subsection{Fine-tuning on Benign Data}
\begin{table*}[]
\small
  \begin{subtable}{0.4\linewidth}
  \centering
    \begin{tabular}{lc}
      \toprule
      \textbf{Model} & \textbf{Biology Accuracy} \\
      \midrule
      Zephyr LLMU     &  0.647         \\
      Zephyr RMU    & 0.599          \\
      \midrule
      Llama LLMU     & 0.713           \\
      Llama RMU     & 0.695           \\
      \bottomrule
    \end{tabular}
    \caption{Biology}
    \label{tab:finetuning biology}
  \end{subtable}
  \quad
  \begin{subtable}{0.6\linewidth}
  \centering
    \begin{tabular}{lcc}
      \toprule
      \textbf{Model} & \textbf{Biology Accuracy} & \textbf{Cyber Accuracy} \\
      \midrule
      Zephyr LLMU     & 0.662  &  0.418           \\
      Zephyr RMU    & 0.657  &  0.411          \\
      \midrule
      Llama LLMU     & 0.727  &   0.471         \\
      Llama RMU     & 0.720  &  0.468          \\
      \bottomrule
    \end{tabular}
    \caption{WMDP}
    \label{tab:finetuning wmdp}
  \end{subtable}
  \caption{Accuracies on the unlearning benchmarks post fine-tuning on Openwebtext.}
  \label{tab:finetuning}
\end{table*}
\noindent The results from fine-tuning on Openwebtext for the Biology dataset are shown in Table~\ref{tab:finetuning}. Fine-tuning is able to recover unlearned performance comparable to (or in some cases even slightly better than) the original model as reported in Table~\ref{tab:robustness tests}. Although the results are from fine-tuning for 1000 steps, in most cases 100 steps are enough to completely recover performance.

\section{Conclusions}

Overall we find RMU tends to perform better than LLMU, particularly in terms of retaining general model capabilities. It has the additional advantage of being less computationally expensive---requiring fewer steps to unlearn and training on fewer layers. LLMU also tends to significantly increase the perplexity of the model. The results from fine-tuning on Openwebtext support the findings of \citet{lucki2024adversarialperspectivemachineunlearning} and extend them to LLMU, and strongly suggest that unlearning methods learn a filter that makes the model refuse to answer harmful queries rather than actually remove information from the model. Our robustness tests show that this filter can be bypassed using simple prompting techniques and motivate the development of unlearning methods that truly remove information from the model. 

\section*{Limitations}

We believe the following points to be the main limitations of our work: (1)~We chose LLMU and RMU as the unlearning methods to experiment on as these were the LLM-specific methods studied by \citet{li2024wmdpbenchmarkmeasuringreducing}. Future work can cover other methods, although experiments by \citet{lucki2024adversarialperspectivemachineunlearning} on NPO \citep{zhang2024negativepreferenceoptimizationcatastrophic} suggest a similar failure at unlearning. (2)~For LLMU, we did not use $\mathcal{L}_{\mathrm{fgt}}$ as our data was not in QA format. Incorporating this loss might have led to improved results. (3)~We observed that training both RMU and LLMU is highly stochastic and that performance varies significantly between checkpoints for even the same set of hyperparameters. Consequently, the results may have been different had we saved checkpoints more frequently. However it is practically infeasible to save the weights more frequently and run all the evaluation tests every time.

\section*{Ethical Considerations}

An adversary could potentially use the techniques introduced in our tests on systems that use these unlearning methods in practice and generate harmful content. However, to our knowledge no system currently makes use of these unlearning methods, and we believe our work promotes safety by detailing the risk of using these methods practically without further modification. 

\FloatBarrier
\bibliography{acl_latex}

\begin{thebibliography}{13}
\providecommand{\natexlab}[1]{#1}

\bibitem[{Dubey et~al.(2024)Dubey, Jauhri, Pandey, Kadian, Al-Dahle, Letman, Mathur, Schelten, Yang, Fan et~al.}]{dubey2024llama3herdmodels}
Abhimanyu Dubey, Abhinav Jauhri, Abhinav Pandey, Abhishek Kadian, Ahmad Al-Dahle, Aiesha Letman, Akhil Mathur, Alan Schelten, Amy Yang, Angela Fan, et~al. 2024.
\newblock \href {https://arxiv.org/abs/2407.21783} {The llama 3 herd of models}.
\newblock \emph{Preprint}, arXiv:2407.21783.

\bibitem[{Henderson et~al.(2023)Henderson, Li, Jurafsky, Hashimoto, Lemley, and Liang}]{JMLR:v24:23-0569}
Peter Henderson, Xuechen Li, Dan Jurafsky, Tatsunori Hashimoto, Mark~A. Lemley, and Percy Liang. 2023.
\newblock \href {http://jmlr.org/papers/v24/23-0569.html} {Foundation models and fair use}.
\newblock \emph{Journal of Machine Learning Research}, 24(400):1--79.

\bibitem[{Hendrycks et~al.(2021)Hendrycks, Burns, Basart, Zou, Mazeika, Song, and Steinhardt}]{hendrycks2021measuringmassivemultitasklanguage}
Dan Hendrycks, Collin Burns, Steven Basart, Andy Zou, Mantas Mazeika, Dawn Song, and Jacob Steinhardt. 2021.
\newblock \href {https://arxiv.org/abs/2009.03300} {Measuring massive multitask language understanding}.
\newblock \emph{Preprint}, arXiv:2009.03300.

\bibitem[{Li et~al.(2024{\natexlab{a}})Li, Pan, Gopal, Yue, Berrios, Gatti, Li, Dombrowski, Goel, Phan et~al.}]{li2024wmdpbenchmarkmeasuringreducing}
Nathaniel Li, Alexander Pan, Anjali Gopal, Summer Yue, Daniel Berrios, Alice Gatti, Justin~D. Li, Ann-Kathrin Dombrowski, Shashwat Goel, Long Phan, et~al. 2024{\natexlab{a}}.
\newblock \href {https://arxiv.org/abs/2403.03218} {The wmdp benchmark: Measuring and reducing malicious use with unlearning}.
\newblock \emph{Preprint}, arXiv:2403.03218.

\bibitem[{Li et~al.(2024{\natexlab{b}})Li, Yong, and Bach}]{li2024preferencetuningtoxicitymitigation}
Xiaochen Li, Zheng-Xin Yong, and Stephen~H. Bach. 2024{\natexlab{b}}.
\newblock \href {https://arxiv.org/abs/2406.16235} {Preference tuning for toxicity mitigation generalizes across languages}.
\newblock \emph{Preprint}, arXiv:2406.16235.

\bibitem[{Liu et~al.(2024)Liu, Yao, Jia, Casper, Baracaldo, Hase, Xu, Yao, Li, Varshney et~al.}]{liu2024rethinking}
Sijia Liu, Yuanshun Yao, Jinghan Jia, Stephen Casper, Nathalie Baracaldo, Peter Hase, Xiaojun Xu, Yuguang Yao, Hang Li, Kush~R Varshney, et~al. 2024.
\newblock Rethinking machine unlearning for large language models.
\newblock \emph{arXiv preprint arXiv:2402.08787}.

\bibitem[{Lynch et~al.(2024)Lynch, Guo, Ewart, Casper, and Hadfield-Menell}]{lynch2024methodsevaluaterobustunlearning}
Aengus Lynch, Phillip Guo, Aidan Ewart, Stephen Casper, and Dylan Hadfield-Menell. 2024.
\newblock \href {https://arxiv.org/abs/2402.16835} {Eight methods to evaluate robust unlearning in llms}.
\newblock \emph{Preprint}, arXiv:2402.16835.

\bibitem[{Tunstall et~al.(2023)Tunstall, Beeching, Lambert, Rajani, Rasul, Belkada, Huang, von Werra, Fourrier, Habib et~al.}]{tunstall2023zephyrdirectdistillationlm}
Lewis Tunstall, Edward Beeching, Nathan Lambert, Nazneen Rajani, Kashif Rasul, Younes Belkada, Shengyi Huang, Leandro von Werra, Clémentine Fourrier, Nathan Habib, et~al. 2023.
\newblock \href {https://arxiv.org/abs/2310.16944} {Zephyr: Direct distillation of lm alignment}.
\newblock \emph{Preprint}, arXiv:2310.16944.

\bibitem[{Weidinger et~al.(2021)Weidinger, Mellor, Rauh, Griffin, Uesato, Huang, Cheng, Glaese, Balle, Kasirzadeh et~al.}]{DBLP:journals/corr/abs-2112-04359}
Laura Weidinger, John Mellor, Maribeth Rauh, Conor Griffin, Jonathan Uesato, Po{-}Sen Huang, Myra Cheng, Mia Glaese, Borja Balle, Atoosa Kasirzadeh, et~al. 2021.
\newblock \href {https://arxiv.org/abs/2112.04359} {Ethical and social risks of harm from language models}.
\newblock \emph{CoRR}, abs/2112.04359.

\bibitem[{Yao et~al.(2023)Yao, Xu, and Liu}]{yao2023large}
Yuanshun Yao, Xiaojun Xu, and Yang Liu. 2023.
\newblock \href {https://openreview.net/forum?id=wKe6jE065x} {Large language model unlearning}.
\newblock In \emph{Socially Responsible Language Modelling Research}.

\bibitem[{Zhang et~al.(2024)Zhang, Lin, Bai, and Mei}]{zhang2024negativepreferenceoptimizationcatastrophic}
Ruiqi Zhang, Licong Lin, Yu~Bai, and Song Mei. 2024.
\newblock \href {https://arxiv.org/abs/2404.05868} {Negative preference optimization: From catastrophic collapse to effective unlearning}.
\newblock \emph{Preprint}, arXiv:2404.05868.

\bibitem[{Zheng et~al.(2024)Zheng, Chiang, Sheng, Zhuang, Wu, Zhuang, Lin, Li, Li, Xing et~al.}]{10.5555/3666122.3668142}
Lianmin Zheng, Wei-Lin Chiang, Ying Sheng, Siyuan Zhuang, Zhanghao Wu, Yonghao Zhuang, Zi~Lin, Zhuohan Li, Dacheng Li, Eric~P. Xing, et~al. 2024.
\newblock Judging llm-as-a-judge with mt-bench and chatbot arena.
\newblock In \emph{Proceedings of the 37th International Conference on Neural Information Processing Systems}, NIPS '23, Red Hook, NY, USA. Curran Associates Inc.

\bibitem[{Łucki et~al.(2024)Łucki, Wei, Huang, Henderson, Tramèr, and Rando}]{lucki2024adversarialperspectivemachineunlearning}
Jakub Łucki, Boyi Wei, Yangsibo Huang, Peter Henderson, Florian Tramèr, and Javier Rando. 2024.
\newblock \href {https://arxiv.org/abs/2409.18025} {An adversarial perspective on machine unlearning for ai safety}.
\newblock \emph{Preprint}, arXiv:2409.18025.

\end{thebibliography}

\appendix



\section{Appendix}

\subsection{Licensing Information}

Wikipedia articles are licensed under CC BY-SA 4.0, i.e. the content is free to use as long as attribution is given. The file names in the dataset are the same as the Wikipedia articles, so they can by used to reference the authors. We release the dataset we create under the same license. The code is released under the MIT license. 

\subsection{Details of Wikipedia Categories Processed} \label{Details of Wikipedia Categories Processed}

\subsubsection*{Unlearning Set}

\textbf{Categories:} Branches of biology, Biological concepts, Eukaryote biology, Biology terminology \\
\textbf{Depth:} 4

\subsubsection*{Retain Set}

\textbf{Categories:} History, Concepts in physics, Philosophy \\
\textbf{Depth:} 3

\subsection{MMLU Biology Subjects} \label{MMLU Biology Subjects}

Anatomy, Clinical Knowledge, College Biology, College Medicine, High School Biology, Human Aging, Medical Genetics, Nutrition, Professional Medicine, Virology
        
\subsection{Sample Zero-shot Question} \label{Sample Zero-shot Question}

\texttt{The following are multiple choice questions (with answers). \\\\
Which of the following structures should be visible on a dental panoramic radiograph? \\
A. The mandibular canal \\
B. The mandibular canal and maxillary sinus \\
C. The mandibular canal, maxillary sinus and hard palate \\
D. The mandibular canal, maxillary sinus, hard palate and zygomatic arch \\
Answer:}

\subsection{Five-shot Prompting}

\subsubsection{List of MMLU Subjects Used}\label{5-shot List}

College Chemistry, College Mathematics, High School Physics, High School World History, Marketing, Miscellaneous, Moral Scenarios, Philosophy, Security Studies, Elementary Mathematics, High School Mathematics, Econometrics, High School Macroeconomics, High School Microeconomics, International Law, Jurisprudence, Professional Law

\subsubsection{Sample Five-shot Question} \label{Sample Five-shot Question}

The subject Management is used here for five-shot prompting \\

\noindent \texttt{The following are multiple choice questions (with answers). \\
What are the two main dimensions of the Ohio Studies into leadership? \\
A. Starting position and end position \\
B. Initial environment and changed environment \\
C. Organisational structure and conditioning \\
D. Initiating structure and considerations \\
Answer: D \\\\
Hygiene factors are associated with which writer? \\
A. Frederick Hertzberg \\
B. D.C. McClelland \\
C. Abraham Maslow \\
D. Douglas McGregor \\
Answer: A \\\\
Which element of the cultural web forms regalia? \\
A. Symbols \\
B. Rituals and routines \\
C. Power structures \\
D. Control systems \\
Answer: A \\\\
What characteristic is not a key feature of the \textquotesingle open systems\textquotesingle\ model of management? \\
A. Morale \\
B. Innovation \\
C. Growth resource \\
D. Adaptation \\
Answer: A \\\\
How can organisational structures that are characterised by democratic and inclusive styles of management be described? \\
A. Hierarchical \\
B. Bureaucratic \\
C. Flat \\
D. Functional \\
Answer: C \\\\
Which of the following structures should be visible on a dental panoramic radiograph? \\
A. The mandibular canal \\
B. The mandibular canal and maxillary sinus \\
C. The mandibular canal, maxillary sinus and hard palate \\
D. The mandibular canal, maxillary sinus, hard palate and zygomatic arch \\
Answer:}

\subsection{Rephrasing Types} \label{Rephrasing Types}

Table~\ref{rephrasing types} contains the names and descriptions of the types of rephrasing done. As an example, the following is the question in Appendix~\ref{Sample Zero-shot Question} rephrased as a poem (the options are not rephrased and so not included): \\

\noindent \texttt{Amidst the dental realm, where shadows play, \\
A panoramic view, a guiding ray, \\
What structures should emerge, clear and defined, \\
To unveil the secrets that our teeth enshrine?}

\begin{table*}[b]
\small
    \begin{tabular}{p{0.3\linewidth} p{0.6\linewidth}}
      \toprule
      \textbf{Name} & \textbf{Description} \\
      \midrule
      Latin Filler Text & Adds Lorem ipsum text before the question \\
      English Filler Text & Adds English text before the question \\
      Hindi Filler Text & Adds Hindi text before the question \\
      Rephrased as Conversation & Rephrases the question as a conversation between two people \\
      Rephrased as Poem & Rephrases the question as a poem \\
      Technical Terms Removed 1 & Replaces technical jargon with simpler vocabulary wherever possible from the question \\
      Technical Terms Removed 2 & Replaces technical jargon with simpler vocabulary wherever possible from the question and answer \\
    Translated to <Language> & Questions and answers translated to <Language> \\
    Replaced With Variables & Some of the terms in the questions are replaced with variables of the form "X", "Y", with these variables being defined before the question \\
    Latin Filler + Rephrased Conversation & Combination of Latin Filler Text and Rephrased as Conversation \\
    English Filler + Rephrased Conversation & Combination of English Filler Text and Rephrased as Conversation \\
    Hindi Filler + Rephrased Conversation & Combination of Hindi Filler Text and Rephrased as Conversation \\
    Latin Filler + Rephrased Poem & Combination of Latin Filler Text and Rephrased as Poem \\
    English Filler + Rephrased Poem & Combination of English Filler Text and Rephrased as Poem \\
    Hindi Filler + Rephrased Poem & Combination of Hindi Filler Text and Rephrased as Poem \\
      \bottomrule
    \end{tabular}
    \caption{Types of rephrasing done on the mutltiple choice questions. Languages used are French, German, Hindi, Korean, Arabic, Czech, Bengali, Vietnamese, Turkish, Telugu and Farsi.}
    \label{rephrasing types}
\end{table*}

\end{document}